\documentclass{article}
\usepackage{blindtext}
\usepackage[a4paper, total={6in, 8in}]{geometry}
\usepackage{natbib}
\setcitestyle{authoryear,open={(},close={)}} 

\usepackage{hyperref}

\usepackage{graphicx}%
\usepackage{multirow}%
\usepackage{amsmath,amssymb,amsfonts}%
\usepackage{amsthm}%
\newtheorem{definition}{Definition}
\usepackage{mathrsfs}%
\usepackage[title]{appendix}%
\usepackage{xcolor}%
\usepackage{textcomp}%
\usepackage{manyfoot}%
\usepackage{booktabs}%
\usepackage{algorithm}%
\usepackage{algorithmicx}%
\usepackage{algpseudocode}%
\usepackage{listings}%
\usepackage{comment}
\usepackage[acronym]{glossaries}
\usepackage{array}
\usepackage{booktabs}
\usepackage{authblk}

\usepackage{subcaption}
\usepackage{xcolor}

\providecommand{\keywords}[1]
{
  \small	
  \textbf{\textit{Keywords---}} #1
} 

\title {Transparency and Proportionality in Post-Processing Algorithmic Bias Correction}

\author[1]{Juliett Suárez Ferreira}
\author[2]{Marija Slavkovik}
\author[3]{Jorge Casillas}
\affil[1]{Data Science and Computational Intelligence Institute (DaSCI), University of Granada.}
\affil[2]{Department of Information Science and Media Studies, University of Bergen}
\affil[3]{Data Science and Computational Intelligence Institute (DaSCI), Department of Computer Science and Artificial Intelligence (DCSAI), University of Granada.}

\date{}

\begin{document}

\maketitle

\begin{abstract}
Algorithmic decision-making systems sometimes produce errors or skewed predictions toward a particular group, leading to unfair results. Debiasing practices, applied at different stages of the development of such systems, occasionally introduce new forms of unfairness or exacerbate existing inequalities. We focus on post-processing techniques that modify algorithmic predictions to achieve fairness in classification tasks, examining the unintended consequences of these interventions. To address this challenge, we develop a set of measures that quantify the disparity in the flips applied to the solution in the post-processing stage. The proposed measures will help practitioners: (1) assess the proportionality of the debiasing strategy used, (2) have transparency to explain the effects of the strategy in each group, and (3) based on those results, analyze the possibility of the use of some other approaches for bias mitigation or to solve the problem. We introduce a methodology for applying the proposed metrics during the post-processing stage and illustrate its practical application through an example. This example demonstrates how analyzing the proportionality of the debiasing strategy complements traditional fairness metrics, providing a deeper perspective to ensure fairer outcomes across all groups. 
\end{abstract}


\section{Introduction}
Bias, as defined by \cite{Tversky1975}, usually signifies a systematic inclination or prejudice that distorts judgment or decision making, causing unfair outcomes. In Artificial Intelligence (AI) systems, bias implies the propensity of a system to consistently produce certain types of error or skewed predictions due to flaws in the data, algorithm design, or training process, and has been recognized as one of the risks of Algorithmic Decision Making (ADM) systems \cite{EUPunderstanding_2019}. For example, a crime prediction model based on historical data collected when crimes of rich people were not recorded would lead to a decision-making algorithm predicting a higher crime proclivity towards poor people \cite{bouchagiar_long_2019}. 

Debiasing refers to a range of strategies and interventions aimed at reducing or eliminating biases in decision-making processes where the goal is to improve objectivity and ensure that decisions are more aligned with normative standards of fairness and accuracy \citep{Tversky1975}. In the development of ADM systems, these interventions have diverse terminology, are called bias mitigation techniques \citep{Siddique2024}, methods for fair machine learning (ML) \citep{Mehrabi2021} or fairness interventions \citep{Caton2024}, and are applied by practitioners at different stages of the development of the ADM system to ameliorate the effect of bias and obtain fairer solutions. These stages, part of the ML pipeline: pre-processing, in-processing, and post-processing, can be observed in Figure \ref{fig:pipe}.

However, an effort to debias a decision can sometimes itself introduce new forms of unfairness or exacerbate existing inequalities. One reason for this is that debiasing techniques may inadvertently privilege certain groups over others in their aim to achieve a fairer result. For example, when adjusting for bias in predictions, post-processing methods can disproportionately impact certain demographic groups by relabeling the outcome of historically advantaged groups to achieve a certain fairness criterion \citep{Barocas2016}. Furthermore, debiasing interventions may not address the root causes of biases but rather shift them in ways that perpetuate or even amplify existing disparities \citep{ONeil2016}. A critical question emerges: {\bf How disproportionate are the results we obtain with the methods we use to debias the outcomes of our algorithms?}

Disproportionality occurs when a debiasing method affects some demographic groups more than others, either by changing their predictions more frequently or by imposing more harmful adjustments (like switching favorable outcomes to unfavorable ones) on one group compared to others. We develop a set of measures to quantify the proportionality of debiasing interventions and define a methodology for applying the proposed metrics in the post-processing stage. 

We focus specifically on examining the unintended consequences of post-processing techniques that modify algorithmic predictions to achieve fairness in classification tasks. Here, we present the study of binary classification and binary protected attributes. However, the proposed metrics can be extended to multi-classification scenarios.

The measures we propose are intended to help practitioners (1) assess the proportionality of the debias strategy used, (2) have transparency that allows them to explain the effects of the strategy in each group, and (3) based on those results, analyze the possibility of using some other strategies for bias mitigation. 

This work is structured as follows. Section \ref{sec:background} reviews some works studying sources of bias in ML and mitigation techniques along its pipeline with a focus on post-processing methods for binary classification tasks. We introduce what proportionality is in this context in Section \ref{sec:proportionality}. Section \ref{sec:metrics} introduces a set of metrics to assess the proportionality of post-processing debiasing interventions, describing their mathematical formulation and characteristics. Section \ref{sec:methodology} presents a methodology for applying the proposed metrics in real-world scenarios along with a practical example of how to use it. This final analysis demonstrates how the proposed metrics provide deeper insights into fairness, complementing traditional fairness metrics with this proportionality analysis. Finally, the discussion, conclusions, and future work highlights the trade-offs identified, discusses the broader implications of proportionality metrics, and identifies opportunities for extending the contributions of this paper.

\section{Background}
\label{sec:background}

Bias exists in many forms, leading to a violation of fairness in some cases. In \citep{Suresh2021, Mehrabi2021, oltenau2016}, the authors discuss various sources of bias in ML, providing categorizations and descriptions to inspire future solutions. Suresh and Guttag \cite{Suresh2021} focus on the presentation of different sources of bias, while Oltenau et al. \cite{oltenau2016} provide an extensive list of different types of bias (along with their definitions) that exist throughout the data lifecycle, from the origin to collection and processing. Mehrabi et al. \cite{Mehrabi2021} incorporate insights from other research papers and introduce a new categorization of these definitions based on data, algorithm and user interaction loop.
 
\begin{figure}[ht]
    \includegraphics[width=0.9\linewidth]{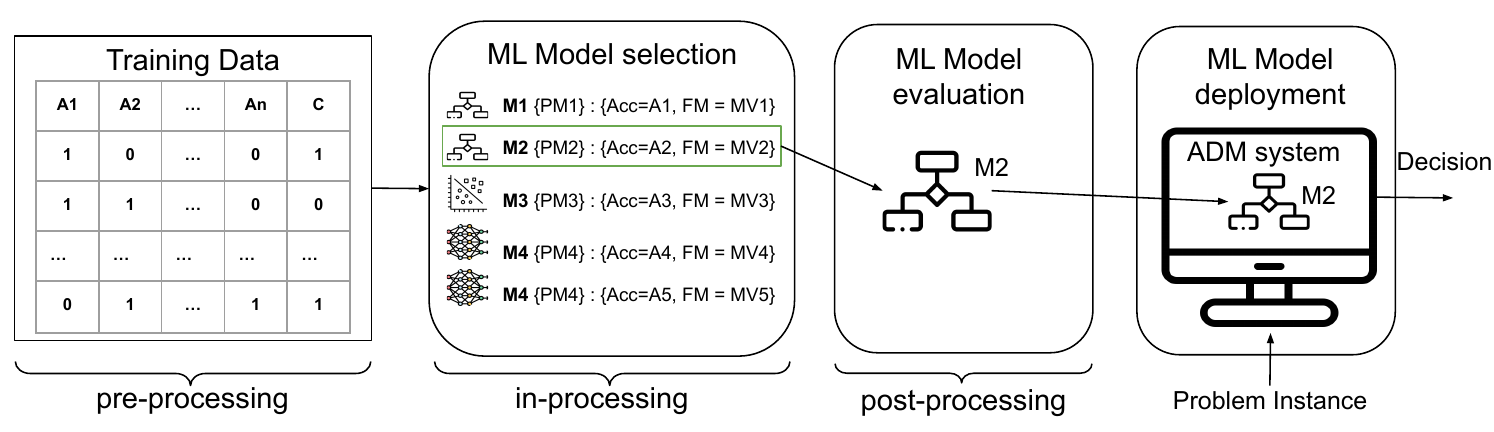}
    \caption{Part of the ML pipeline highlighting pre-processing, in-processing and post-processing interventions at different stages as well as the ADM process.}
    \label{fig:pipe}
\end{figure}
 
A variety of mitigation techniques have been developed, each targeting different stages of the ML pipeline where bias can manifest. They are categorized broadly into three types: pre-processing, in-processing, and post-processing approaches \citep{Mehrabi2021, Caton2024, Siddique2024}. Each category represents different stages of the ML pipeline where practitioners can apply different interventions to mitigate bias, and each has distinct operational mechanisms and implications on the outcomes, see Figure~\ref{fig:pipe}.

We focus on the results of the post-processing methods specifically designed for classification tasks where the goal is to predict a label (y) from a set of inputs using a pre-trained model\footnote{A model can be considered a mathematical function that map input data to output predictions. For classification tasks, they are produced by training an algorithm with predefined data and using specific parameters.}. The debiasing methods in these cases are applied after the model has been trained and act by modifying the prediction of the model to ensure fairness without altering the model itself or the training data. The main techniques include calibration \citep{Pleiss2017}, thresholding \citep{Kamiran2012b} and transformation \citep{nabi18fair}.

\begin{figure}[ht]
    \centering
    \includegraphics[width=0.9\linewidth]{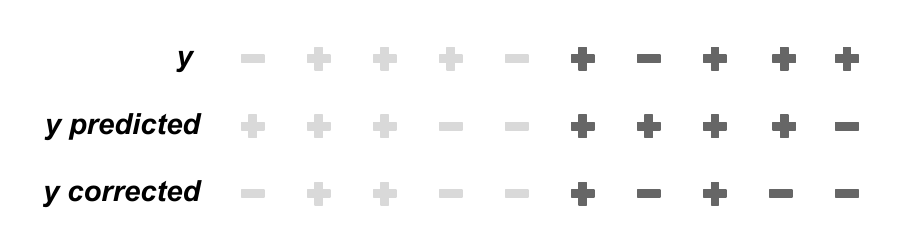}
    \caption{A post-processing debiasing method that works by changing the label of the instances. Consider \textit{\textbf{y}} the label of the instance, \textit{\textbf{y predicted}} the predicted label and \textit{\textbf{y corrected} }the label obtained after applying a post-processing debias strategy. The signs, \textbf{$+$} and \textbf{$-$ }represent the value of the class and the colors (light and dark gray) indicate the membership of each instance to a particular group taking into account a protected attribute such as gender, race, etc.}
    \label{fig:debiasing}
\end{figure}

We illustrate a very simple case in Figure \ref{fig:debiasing}. The figure represents the outcome of 10 instances of a classification problem (\textit{y}) with two possible labels (+ and -) belonging to two groups (light gray and dark gray). Labels represent the outcome of the classification and groups symbolize a protected attribute. The predicted labels for the dark gray group (\textit{y predicted}) have 4 out of 5 examples in the positive class (80\%) while the pink group has 3 out of 5 examples in the positive class (60\%) this will represent a difference of 20\% in statistical parity\footnote{A fairness metric which declares that the likelihood of a positive outcome should be the same irrespective of an individual's group membership.} \citep{Calders2009} which could be considered unfair. Consider that applying a post-processing debiasing method obtains the \textit{y corrected} labels. Although the number of positive outcomes in the two groups has the same proportion (2 out of 5 for a 40\%), we can observe that even when the flips occur in both groups towards the negative label, they impact more the dark gray group compared to the light gray group. Here, we aim to evaluate the unintended consequences of debiasing techniques that result in the alteration of the outcomes with the purpose of achieving fairness.

\section{What Proportionality Means}
\label{sec:proportionality}

Classic group fairness metrics (e.g. statistical parity or equalized odds) summarize \emph{disparities} in the final predictions produced by a classifier. They do not, however, reveal \emph{how} a post-processing intervention arrived at those predictions, nor who gained or lost during that process.  
\textbf{Proportionality} fills this gap: it asks whether the \emph{benefits} and \emph{burdens} that arise when we \emph{flip} labels in the post-processing stage are distributed in a way that is \emph{normatively justified} and \emph{legally defensible}. 

Let a \emph{flip} be any change of a predicted label induced by a post-processing rule. We distinguish:

\begin{itemize}
    \item \textbf{Balancing flips across groups.}  The counts and rates of flips should not be so unequal that one group bears virtually all harmful flips or garners all beneficial ones.  Our metrics (Section~\ref{sec:metrics}) make this distribution explicit, extending the logic of statistical parity \citep{Calders2009} from \emph{outcomes} to \emph{interventions}.

    \item \textbf{Harmful versus beneficial flips.}  
          Changing a positive outcome to a negative one is usually a genuine loss for the affected individual (e.g., losing a job offer or loan).  
          A proportional rule therefore seeks to \emph{minimise} harmful flips overall and to avoid concentrating them on historically marginalised groups.  
          By separately tracking harmful and beneficial flips, we expose potential levelling-down (many losses for one group, few gains elsewhere) and encourage \emph{levelling-up} strategies that improve outcomes for disadvantaged groups \citep{Weerts2022}.
\end{itemize}

Numerous dimensions of moral and political philosophy elucidate the significance of proportionality. The equality of opportunities requires that people with comparable talent or qualifications face similar chances of desirable outcomes, regardless of protected attributes \citep{Fishkin2014,Khan2022}. A post-processing rule that places most negative flips on one group violates this principle, whereas a rule \emph{proportionate} levels the playing field without arbitrarily closing doors to the otherwise qualified.

Desert-based accounts hold that benefits should align with effort or qualification \citep{Miller1999}. Excessive harmful flips against high performers signal a desert violation; a proportionate intervention corrects bias while continuing to reward merit.

Sufficiency and prioritarian theories prioritize improving the situation of those who are worse off \citep{Arneson2000}. Therefore, proportionality disfavors \emph{levelling down}, making advantaged groups worse off without materially helping the disadvantaged, a practice strongly criticized \citep{Mittelstadt2023}.

Proportionality also echoes the established equality doctrine. \textit{EU fundamental-rights law} applies a four-step proportionality test (suitability, necessity, balancing, and consistency \citep{EUProportionalityGuidelines}) whenever a policy imposes differential treatment \citep{EUCharter}.  \textit{The UK Equality Act} adopts a near-identical standard for justifying indirect discrimination: the measure must be ``a proportionate means of achieving a legitimate aim'' \citep{EqualityAct2010}.  Our proposal operationalize these legal ideas: they quantify whether a debiasing strategy imposes an \emph{excessive} share of negative flips on any group and thus provide empirical evidence for (or against) legal proportionality.

Proportionality evaluates \emph{whether fairness corrections themselves are fair}. Grounding our metrics in normative theory and equality law serves to equip practitioners with principled diagnostics that complement traditional group-fairness statistics and guard against well-intentioned but excessive interventions.

\section{Assessing the Effects of Flips Produced by Post-Processing Debiasing Techniques}
\label{sec:metrics}

In this section, we provide a characterization of the flips in the solution (Section \ref{sec_characterization}) that offer a general picture of how the debiasing algorithm affects the model's predictions, and subsequently, we extend the analysis with the objective of evaluating whether the debiasing algorithm impacts different groups equitably by introducing a series of flip proportionality metrics in Section \ref{sec:FPMetrics}. 

For each metric, we present both its mathematical definition and its interpretation. Each subsection concludes with a summary table of the proposed metrics, clarifying their boundaries, a short description, and edge cases. 

\subsection{Characterization of Flips in a Solution}
\label{sec_characterization}

In this Section we define key concepts and metrics that quantify how a debiasing algorithm modifies the original predictions. 

We first start by characterizing a classification problem: given a set of features $ X \in \mathbb{R}^{n \times d}$, where $ n $ is the number of instances and $ d $ is the number of features. The goal of the classification task is to learn a classifier $ f: \mathbb{R}^d \rightarrow {0, 1} $ that predicts the binary outcome $ y \in {0, 1}^n : \hat{y} = f(X)$ where $ X = \{x_1, x_2, \dots, x_n\} $ represents the feature vectors for $ n $, $ y = \{y_1, y_2, \dots, y_n\} $ represents the true binary outcomes, and $ \hat{y} = \{\hat{y}_1, \hat{y}_2, \dots, \hat{y}_n\} $ represents the predicted outcomes.

Let \( y_{\text{predicted}} = \hat{y} \in \{0, 1\}^n \) be the vector of predicted labels from the classifier for the same instances \( n \) and consider that $0$ is the unfavorable outcome and $1$ the favorable one. For illustrative purposes, consider a classification problem in which possible outcomes entail either accepting or rejecting a candidate. A favorable or positive outcome is accepting the candidate and will have the value $1$ in $y_{\text{predicted}}$.

\begin{equation}
y_{\text{predicted}} = \{y_{\text{predicted},1}, y_{\text{predicted},2}, \dots, y_{\text{predicted},n}\}
\end{equation}

After applying a debiasing algorithm, the predicted labels can be adjusted to form the corrected labels \( y_{\text{corrected}} \in \{0, 1\}^n \).

\begin{equation}
y_{\text{corrected}} = \{y_{\text{corrected},1}, y_{\text{corrected},2}, \dots, y_{\text{corrected},n}\}
\end{equation}

We will compare $y_{predicted}$ with $y_{corrected}$. This comparison isolates the effect of the debiasing algorithm on the classifier's predictions ($y_{predicted}$) measuring how much the debiasing method has adjusted the predictions to correct potential biases. A flip occurs when the label changes from one label to another (e.g., from 0 to 1 or from 1 to 0) as a result of a debias algorithm. 

\begin{definition}[Flip]
Let \( y_{\text{predicted}} \) and \( y_{\text{corrected}} \)  be sets of the corresponding outputs after a debiasing process. 

A flip between \( y_{\text{predicted}} \) and \( y_{\text{corrected}} \) is defined as:

\begin{equation}
\text{Flip}_i = 
\begin{cases} 
1 & \text{if } y_{\text{predicted},i} \neq y_{\text{corrected},i} \\
0 & \text{if } y_{\text{predicted},i} = y_{\text{corrected},i} 
\end{cases}
\end{equation}
\end{definition}

\begin{definition}[Number of flips]
The number of flips, $N_{\text{flips}}$, represents the total count of instances in which the predicted label $y_{\text{predicted}}$ differs from the corrected label $y_{\text{corrected}}$ after applying the debiasing algorithm:

\begin{equation}
N_{\text{flips}} = \sum_{i=1}^{n} \text{Flip}_i
\label{eq_nflips}
\end{equation} 
\end{definition}

\begin{definition}[Flip Rate (\textbf{FR})] Is defined as the proportion of instances that experienced a flip over the total number of instances. 

\begin{equation}
\text{Flip Rate} = \frac{N_{\text{flips}}}{n}
\label{eq_fliprate}
\end{equation}

Where $N_{\text{flips}}$ is the number of flips defined in Equation \ref{eq_nflips} and $n$ is the total number of instances.
\end{definition}

When this proportion constitutes a substantial percentage of the overall number of instances, demonstrate that the post-processing algorithm has introduced significant modifications to the algorithm's output, indicating potential fairness issues within the applied algorithm. Practitioners in those cases should consider the possibility of modifying or improving the algorithm applied with the aim of minimize the post-processing interventions.

In the context of evaluating the impact of debiasing algorithms, we classify flips based on their effect on the predicted outcomes. Specifically, we distinguish between flips that lead to a positive change (from unfavorable to favorable values) and negative changes (from favorable to unfavorable values) in the predicted labels. Elaborating on the scenario of accepting or rejecting a candidate; a transition towards a positive outcome would mean a change in prediction from rejection to acceptance.

\begin{definition}[Favorable Flips] A positive flip that occurs when the predicted label changes from unfavorable outcome ($0$) to a favorable outcome ($1$), indicating an increase in the number of positive decisions. This type of flip represents a shift towards a more favorable outcome for the instance.

\begin{equation}
\text{Favorable Flip}_i =
\begin{cases} 
1 & \text{if } y_{\text{predicted},i} = 0 \text{ and } y_{\text{corrected},i} = 1 \\
0 & \text{otherwise} 
\end{cases}
\label{eq_favflip}
\end{equation}

The total number of positive flips, \( N_{\text{favorable flips}} \), is given by:

\begin{equation}
N_{\text{favorable flips}} = \sum_{i=1}^{n} \text{Favorable Flip}_i
\label{eq_totfavflips}
\end{equation}
\end{definition}

\begin{definition}[Unfavorable Flips] A negative flip or a harmful flip occurs when the predicted label changes from $1$ to $0$, indicating a decrease in the number of positive decisions. This type of flip represents a shift towards a less favorable outcome for the instance.

\begin{equation}
\text{Unfavorable Flip}_i =
\begin{cases} 
1 & \text{if } y_{\text{predicted},i} = 1 \text{ and } y_{\text{corrected},i} = 0 \\
0 & \text{otherwise} 
\end{cases}
\label{eq_unfavflip}
\end{equation}

The total number of negative flips, \( N_{\text{unfavorable flips}} \), is given by:

\begin{equation}
N_{\text{unfavorable flips}} = \sum_{i=1}^{n} \text{Unfavorable Flip}_i
\label{eq_totunfavflips}
\end{equation}

\end{definition}

These classifications help in understanding the nature of the changes made by the debiasing algorithm. Analyzing the nature of the flips can provide insight into how the debiasing process impacts overall decision making.

\begin{definition}[Directional Flip Ratio (\textbf{DFR})] Compares the number of favorable flips (from $0$ to $1$) with unfavorable flips (from $1$ to $0$).

    \begin{equation}
    \text{DFR} = \frac{N_{\text{favorable flips}}}{N_{\text{unfavorable flips}}}
    \label{eq_dfr}
    \end{equation}

\end{definition}

A DFR closer to $1$ indicates balanced flip directions, suggesting that the debiasing algorithm is not disproportionately flipping predictions in one direction (e.g., systematically downgrading or upgrading individuals).
    
Values greater than $1$ suggest more favorable than unfavorable flips, and values lower than $1$ will suggest the higher occurrence of unfavorable flips. The desired values for this metric should be close to 1 indicating balanced flips in both directions. 

Taking into account unfavorable flips (those in which the outcome was changed to an unfavorable value), we can establish metrics of the impact on individuals when achieving fairness. These flips are considered harmful because they represent a tangible loss for the affected individuals; in the previous example, when a prediction changes from job candidate acceptance to rejection. While such changes may be necessary to achieve overall system fairness, they represent real negative consequences for the individuals whose predictions are flipped, making it crucial to measure and minimize their occurrence, especially when they disproportionately affect specific demographic groups.

\begin{definition}[Harmful Flip Proportion(\textbf{HFP})] This metric calculates the proportion of harmful flips among all flips. The HFP is then defined as:

    \begin{equation}
    \text{HFP} = \frac{N_{\text{unfavorable flips}}}{N_{\text{flips}}}
    \label{eq_HFP}
    \end{equation}

Where $N_{\text{unfavorable flips}}$ was defined in Equation \ref{eq_unfavflip} and $N_{\text{flips}}$ is the total number of flips defined in Equation \ref{eq_nflips}.

\end{definition}

A harmful flip was defined as a change in prediction from a positive to a negative outcome, which is interpreted as a detrimental change for the individual instance. A lower HFR indicates that fewer flips result in harmful outcomes, suggesting that the debiasing algorithm is less likely to produce negative effects on the predictions. 

The metrics introduced until now allow us to characterize the flips made by a debiasing algorithm that transforms the output in the post-processing stage. After their calculations, practitioners will have a general overview of the flips applied to the solution. Furthermore, these metrics can be independently applied to each distinct group, providing practitioners with an understanding of the overall incidence of flips in each group. A resume of the metrics proposed to describe the flips in the solution is detailed in Table \ref{tab:sumary_solution} of the Appendix \ref{summary}.

\subsection{Group-Based Flip Proportionality Metrics}
\label{sec:FPMetrics}

In this section we propose flip proportionality metrics to calculate the differences between the flips inflicted on the groups. For group-based metrics, let us consider that instances are characterized by their belonging to a certain binary protected feature, where $1$ represents individuals with historical advantage (privileged) and $0$ represents individuals in historical disadvantage (unprivileged). The Flip Rate for each group can be calculated separately.

Let us define $ S \in \{0, 1\}^n $ as the vector indicating the membership of the protected group for each instance, where \( S_i = 1 \) indicates a privileged individual and \( S_i = 0 \) indicates an unprivileged individual.

The Flip Rate for the privileged group (\( \text{FR}_{\text{privileged}} \)) and the unprivileged group (\( \text{FR}_{\text{unprivileged}} \)) can be formulated as:

\begin{subequations}
\begin{equation}
\text{FR}_{\text{privileged}} = \frac{\sum_{i=1}^{n} \text{Flip}_i \cdot \mathbb{I}(S_i = 1)}{\sum_{i=1}^{n} \mathbb{I}(S_i = 1)}
\end{equation}

\begin{equation}
\text{FR}_{\text{unprivileged}} = \frac{\sum_{i=1}^{n} \text{Flip}_i \cdot \mathbb{I}(S_i = 0)}{\sum_{i=1}^{n} \mathbb{I}(S_i = 0)}
\end{equation}
\label{eq_FRG}
\end{subequations}

Where \( \mathbb{I}(S_i = 1) \) is an indicator function equal to 1 if the instance \( i \) belongs to the privileged group, and 0 otherwise and \( \mathbb{I}(S_i = 0) \) is an indicator function equal to 1 if the instance \( i \) belongs to the unprivileged group, and 0 otherwise.

The same way, the metric Harmful Flip Proportion can be calculated separately for different groups. For instance, \( \text{HFP}_{\text{unprivileged}} \) represents the HFP for the unprivileged group, and \( \text{HFP}_{\text{privileged}} \) represents the HFP for the privileged group:

\begin{subequations}
    \begin{equation}
        \text{HFP}_{\text{privileged}} = \frac{\sum_{i=1}^{n} \text{Unfavorable Flip}_i \cdot \mathbb{I}(S_i = 1)}{\sum_{i=1}^{n} \text{Flip}_i \cdot \mathbb{I}(S_i = 1)}
    \end{equation}
        
    \begin{equation}
        \text{HFP}_{\text{unprivileged}} = \frac{\sum_{i=1}^{n} \text{Unfavorable Flip}_i \cdot \mathbb{I}(S_i = 0)}{\sum_{i=1}^{n} \text{Flip}_i \cdot \mathbb{I}(S_i = 0)}
    \end{equation}
    \label{eq_HFPG}
\end{subequations}

Taking the definition of flips rates and harmful flip proportions for each group, we define a set of proportionality measures that quantify the disparity in the flips between the groups.
    
\begin{definition}[Flip Rate Difference (\textbf{FRD}) \& Harmful Flip Proportion Difference (\textbf{HFPD})] This metric calculates the absolute difference in the flip rates or the harmful flip rates between two groups: 

\begin{subequations}
    \begin{equation}
    \text{FRD} = \left|\text{FR}_{\text{privileged}} - \text{FR}_{\text{unprivileged}}\right|
    \label{eq_FRD}
    \end{equation}  
    \begin{equation}
    \text{HFPD} = \left|\text{HFP}_{\text{privileged}} - \text{HFP}_{\text{unprivileged}}\right|
    \label{eq_HFPD}
    \end{equation}
    \label{eq_FRD_HFPD}
\end{subequations}

\end{definition}

A value close to $0$ for FRD or HFPD indicates a more proportional treatment between groups, so the desirable value is zero.

These metrics capture the similarities between the flip rates and the proportion of harmful flips between the groups. However, if the size of one group is significantly smaller, these metrics might overemphasize disparities due to variance in smaller sample sizes. Therefore, they should be accompanied by the analysis of the individual measures for both groups independently defined in Equations \ref{eq_FRG} and \ref{eq_HFPG}, as it will capture the proportions between the flip rates and the proportion of harmful flips of the groups separately.  This distinction allows for the identification of scenarios that can have small differences but high values separately that suggest a high incidence of flips in the overall solution.

\begin{definition}[Disparity Index (\textbf{DI}) \& Harmful Disparity Index (\textbf{HDI})] The disparity index highlights the disparity between the flip rates or harmful flip proportions of two groups. 

\begin{subequations}
    \begin{equation}
    \text{DI} = \frac{\max(\text{FR}_{\text{privileged}}, \text{FR}_{\text{unprivileged}})}{\min(\text{FR}_{\text{privileged}}, \text{FR}_{\text{unprivileged}})}
    \label{eq_DI}
    \end{equation}
    \begin{equation}
    \text{HDI} = \frac{\max(\text{HFP}_{\text{privileged}}, \text{HFP}_{\text{unprivileged}})}{\min(\text{HFP}_{\text{privileged}}, \text{HFP}_{\text{unprivileged}})}
    \label{eq_HDI}
    \end{equation}
    \label{eq_DI_HDI}
\end{subequations}
\end{definition}

A DI or HDI of $1$ indicates perfect proportionality, while values greater than $1$ indicate the extent of disproportionality. These metrics use ratios, and if the denominator (flip rate or harmful flip proportion) is very small, the disparity index can become disproportionately large, particularly for groups with fewer flips. Therefore, an analysis in conjunction with the characterization of the flips is required to better understand the overall flip context. 

\begin{definition}[Flip Rate Disparity (\textbf{FD}) \& Harmful Flip Proportionality Disparity (\textbf{HFD})] This metric computes the difference in the flip rates or harmful flip proportion between different groups in relation to the overall \text{Flip Rate} defined in Equation \ref{eq_fliprate}. 

\begin{subequations}
    \begin{equation}
        \text{FD} = \left| \frac{\text{FR}_{\text{privileged}}}{\text{Flip Rate}} - \frac{\text{FR}_{\text{unprivileged}}}{\text{Flip Rate}} \right|
    \label{eq_FD}
    \end{equation}

    \begin{equation}
        \text{HFD} = \left| \frac{\text{HFP}_{\text{privileged}}}{\text{Flip Rate}} - \frac{\text{HFP}_{\text{unprivileged}}}{\text{Flip Rate}} \right|
    \label{eq_HFD}
    \end{equation}
    \label{eq_FD_HFD}
\end{subequations}
\end{definition}

If the overall flip rate or the harmful flip proportion is close to $0$, the normalized rates might become very large, potentially magnifying the value of the measures. This should be taken into account for better interpretability of the measure; also these measures do not explicitly account for group sizes. If one group (e.g., the privileged group) is much smaller, its flip rate might have a higher variance, potentially skewing the value of the measures. Values close to $0$ are the desirable values for these measures. 

\begin{definition}[Relative Flip Disparity (\textbf{RFD}) \& Relative Harmful Flip Disparity (\textbf{RHFD})] These metrics provide a normalized measure of the disparity between the flip rates or the harmful flip proportions between groups. 

\begin{subequations}
    \begin{equation}    
        \text{RFD} = \frac{FRD}{\text{FR}_{\text{unprivileged}} + \text{FR}_{\text{privileged}} }
        \label{eq_RFD}
    \end{equation}
    \begin{equation}    
        \text{RHFD} = \frac{HFPD}{\text{HFP}_{\text{unprivileged}} + \text{HFP}_{\text{privileged}} }
        \label{eq_RHFD}
    \end{equation}
    \label{eq_RFD_RHFD}
\end{subequations}

Where the values of FRD and HFPD are defined in equations \ref{eq_FRD} and \ref{eq_HFPD} respectively.

\end{definition}

These metrics are beneficial for understanding the relative difference in flip rates (or harmful flip rates) in a way that is proportionate and comparable across different scenarios. A value closer to $0$ indicates that the flip rates or harmful flip proportions between the groups are proportionally similar, implying that the debiasing process similarly affects both groups. Higher values of these measures indicate a greater disparity between the groups, suggesting that one group is experiencing flips at a significantly different rate than the other.

These metrics normalize disparities based on the sum of group-specific rates. If the overall rates are small or one group dominates, normalized metrics might amplify the disparities. For clarity, we give a summary of the proposed metrics in Table~\ref{tab:sumary_group} of the Appendix \ref{summary}.

\section{Applying the Proposed Metrics: Methodology and Example}
\label{sec:methodology}

\begin{figure}[ht]
    \centering
    \includegraphics[width=0.99\linewidth]{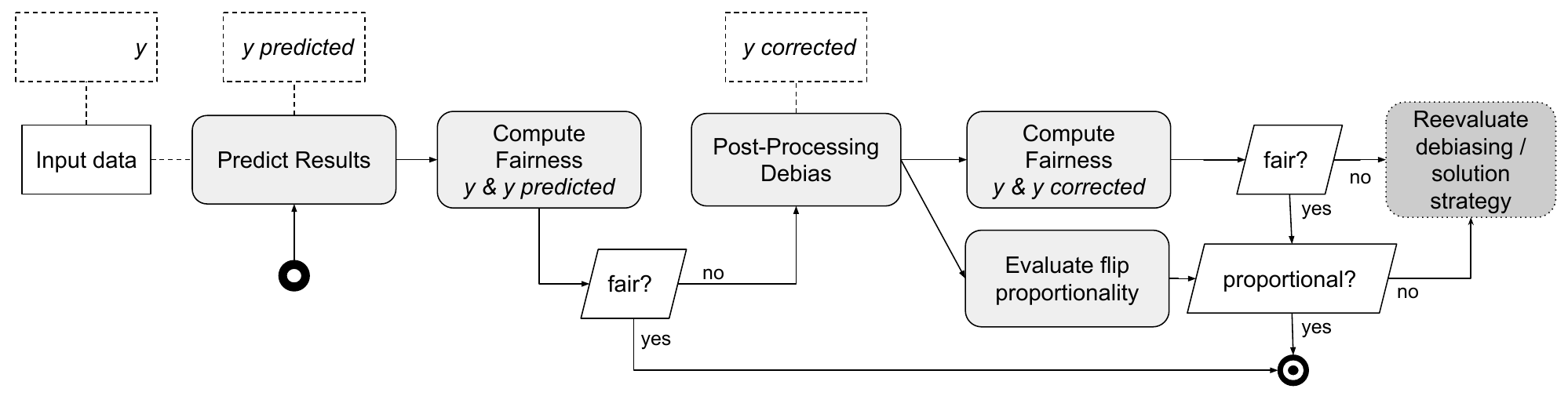}
    \caption{Methodology}
    \label{fig:methodology}
\end{figure}

In this Section, we discuss how to apply the measures we defined,  while analyzing the results of a post-processing dibiasing method that works by flipping the output of the solution. Figure \ref{fig:methodology} illustrates a step-by-step debiasing strategy to achieve fairness in the predictions while evaluating proportionality. The process begins by computing the predicted labels $y_{predicted}$. The fairness of these predictions is evaluated by comparing the true ($y$) and predicted labels. If fairness criteria are not met, a debiasing post-processing step is applied, producing corrected labels ($y_{corrected}$). Then these corrected labels are again evaluated for fairness. Furthermore, the method assesses whether the correction of the predicted values is proportional, ensuring that the changes applied do not disproportionately affect specific groups. The analysis ends when both fairness and proportionality are achieved; otherwise, practitioners should evaluate whether to change the debiasing strategy or the solution. It is important to note that any group fairness metric can be used appropriately to evaluate classification outcomes.

We demonstrate the application of the proposed methodology through a specific example. The results and algorithms in this section are not intended as contributions to this paper, but rather they serve to demonstrate the application of the proposed proportionality metrics.

First, we solved a toy problem using a \textit{DecisionTreeClassifier} implemented in the Scikit-learn library \citep{scikit-learn}; the accuracy of the classification is $0.725$. Then, we calculated two fairness metrics for the solution: Statistical Parity (SP) \citep{Calders2009} and Equalized Odds (EO) \cite{Hardt2016} using the \emph{AI Fairness 360} (AIF360) toolkit \cite{Bellamy}. The results of these metrics were $-0.31$ and $0.28$ respectively which indicate the disparity between the privileged and unprivileged groups. After that, we applied the \textit{EqOddsPostprocessing} algorithm based on \cite{Hardt2016}. The results of SP and EO after the post-processing step are $-0.071$ and $0.025$, respectively, which are accepted values in the fair interval $[-0.1, 0.1]$.  

\begin{figure}
    \centering
    \includegraphics[width=0.9\linewidth]{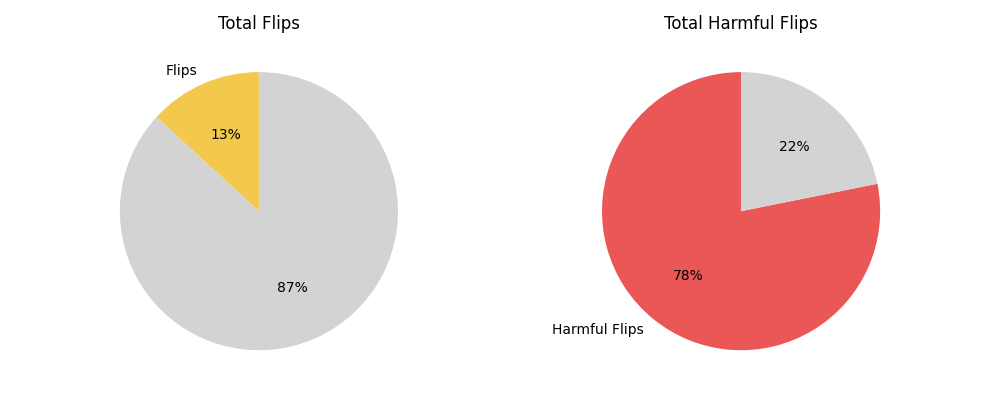}
    \includegraphics[width=0.9\linewidth]{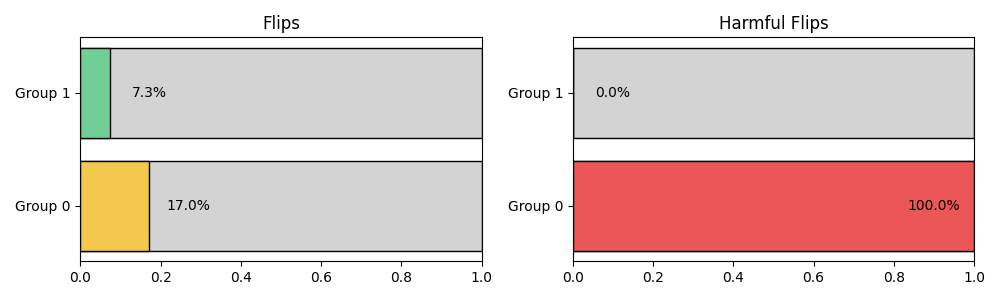}
    \includegraphics[width=0.9\linewidth]{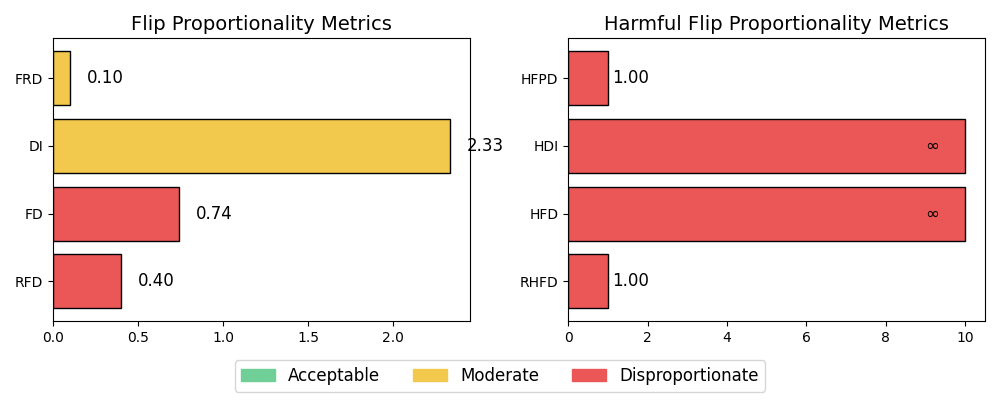}
    \caption{Visual analysis of Flips and Group-Based Flip Proportionality Metrics}
    \label{fig:flip-viz}
\end{figure}

The first interpretation is that the post-processing method has solved the fairness problem. Nevertheless, when a closer look is taken to the flips occurring in the debias process, the appropriateness of the results may be revised. To achieve this, we have implemented the proposed metrics in Python\footnote{The implementation code can be provided as supplementary material or published in a public repository}, as output we offer a proportionality report with the results of the measures as is illustrated in Table \ref{tab:exampleresults} and a visualization of the main metrics as observed in Figure \ref{fig:flip-viz}. 

We used the predicted and corrected results of our toy problem and show the results of the metrics applied to the example in Table \ref{tab:exampleresults}. The table lists the metric name (values in bold in the column \textit{Metric}), their value (\textit{Result}) and a \textit{Short Analysis} of the calculation. This short analysis is also produced by the code implemented based on the computation of the metric. The table also informs about the dataset, groups, and flipped totals.  

\begin{longtable}{llp{6cm}}
\caption{Result of the metrics characterizing the flips and the flip proportionality metrics of the toy problem used as an example.}
\label{tab:exampleresults}
\\
\hline
\textbf{Metric} & \textbf{Result} &  \textbf{Short Analysis} \\ \hline
\endfirsthead
\hline
\textbf{Metric} & \textbf{Result} &  \textbf{Short Analysis}  \\ \hline
\endhead

\multicolumn{3}{l}{Dataset information} \\
\midrule
Total samples & 1320 & \\
Group 0 samples & 799 & \\
Group 1 samples & 521 & \\
\midrule

\multicolumn{3}{l}{Overall Metrics} \\
\midrule
Total flips& 174& \\
\textbf{FR} & 0.13 & \\
Harmful Flips & 136 & \\
\textbf{HFP} & 0.78 & Regular calculation \\
\midrule

\multicolumn{3}{l}{Flips by Groups} \\
\midrule
Group 0 Flips & 136 & \\
Group 0 \textbf{FR} & 0.17 & \\
Group 0 Harmful flips & 136 & \\
Group 0 \textbf{HFP} & 1.0 & Only harmful flips \\
Group 1 Flips & 38 & \\
Group 1 \textbf{FR} & 0.073 & \\
Group 1 Harmful flips & 0.0 & \\
Group 1 \textbf{HFP} & 0.0 & No harmful flips \\
\midrule

\multicolumn{3}{l}{Directional flip ratio} \\
\midrule
\textbf{DFR} & 0.28 & Regular calculation \\
Group 0 \textbf{DFR} & 0.0 & Only harmful flips \\ 
Group 1 \textbf{DFR} & $\infty$ & Only beneficial flips \\
\midrule

\multicolumn{3}{l}{Flip Proportionality Metrics} \\
\midrule
\textbf{FRD} & 0.097 & \\
\textbf{DI} & 2.33 & Regular calculation \\
\textbf{FD} & 0.74 & Regular calculation \\
\textbf{NFD} & 0.40 & Regular calculation \\
\midrule

\multicolumn{3}{l}{Harmful Flip Proportionality Metrics} \\
\midrule
\textbf{HFRD} & 1.0 & \\
\textbf{HDI} & $\infty$ & One value is zero\\
\textbf{HFD} & $\infty$ & One value is zero \\
\textbf{NHFD} & 1.0 & Regular calculation \\
\bottomrule
\end{longtable}

When we analyze the values in Table \ref{tab:exampleresults}, we observe that while the flip rate value for the overall dataset is $13\%$ the harmful flip proportion constitutes $78\%$ of the flipped instances, raising concerns about the fairness of the process generating the flips. Analyzing the flips in each group reveals that most of them occur in Group 0, and all flips in this group are harmful. In contrast, Group 1 experiences fewer flips, all of which are beneficial. The directional flip rate highlights the disparity between the groups in terms of the nature of their flips.

Collective analysis of flip proportionality metrics demonstrates a systematic disparity in flip rates between the groups. This issue is exacerbated when examining harmful flip proportionality metrics, which expose significant fairness disparities between the groups, highlighted by the concentration of all harmful flips within Group 0.

We have commented on the desirable values for each proportionality metric, but have not proposed thresholds in which the results of the metrics can be considered acceptable, moderate, or disproportionate. We consider that these thresholds will also depend on the context of the problem analyzed since factors such as the size of the dataset may influence the metrics values. 

In our implementation of the proposed metrics, we provide an example of threshold values that can be configured. We consider that a difference lower than $0.1$ from each proportionality metric result with respect to their ideal value can be considered acceptable, a difference in a range of $[0.1, 0.3]$ indicate manageable disparities requiring review, and beyond that difference, we consider the values to show imbalance that may indicate problems with the proportionality of the debiasing strategy applied. We applied these ranges for all the metrics except by the FRD \& HFPD (Equation \ref{eq_FRD_HFPD}) in which we consider a difference of $0.05$ for acceptable values and between $0.05$ and $0.15$ for moderate values.

We encourage practitioners to consider the adaptation of these values to their particular problem. The proposed thresholds are empirical and are not mathematically derived or rigorously proved; instead, they are based on practical considerations to guide the interpretation of results. 

For an easier interpretation of the results, we have implemented a simple visualization of the main metrics. The visual analysis can be seen in Figure \ref{fig:flip-viz}. The figure presents a visual analysis of flips and flip proportionality measures. The upper graph displays the Flip Rate and the Harmful Flip Proportion as percentages. The middle graph presents the same metrics per group. The lower graph focuses on flip proportionality metrics. Each metric is color-coded according to the thresholds explained previously as Acceptable (green), Moderate (yellow), and Disproportionate (red). We also limit the $\infty$ values to a maximum for better visualization. 

The analysis of Figure \ref{fig:flip-viz} reinforces the results already analyzed from Table \ref{tab:exampleresults}. As an inference of this toy problem, in real-world scenarios, practitioners should analyze the suitability of the debiasing strategy, as well as the possibility of applying other methods to solve the problem; otherwise, it is necessary to provide a justification for the disproportionate adverse treatment experienced by a specific group. The example was instrumental in illustrating the unintended consequences of debiasing strategies, particularly in terms of the harm experienced by a particular group that is masked within the fairness metrics results.

\section{Discussion and Limitations}

Our analysis of proportionality metrics underscores crucial trade-offs inherent in algorithmic debiasing, highlighting the complexity of ensuring equitable impacts across demographic groups. Although disproportional flipping might sometimes be unavoidable due to underlying data distributions or historical biases, practitioners should transparently document and ethically justify such occurrences. Furthermore, alternative debiasing strategies, such as pre-processing or in-processing methods, should be actively explored to achieve fairness without disproportionate impacts.

We recognize that evaluating proportionality solely through prediction flips does not fully capture the nuanced interplay between fairness interventions, predictive accuracy, and ground truth labels. Some flips initiated to improve fairness may incidentally align predictions with actual outcomes, thus improving accuracy; conversely, others may inadvertently degrade predictive quality. Therefore, integrating proportionality metrics alongside traditional accuracy indicators (e.g., precision and recall) and fairness measures is essential. This evaluation allows practitioners to better differentiate between beneficial corrections and fairness-driven errors, facilitating more informed and ethically sound decision-making.

Despite their utility, proportionality metrics exhibit certain limitations. Specifically, the metrics may be overly sensitive in scenarios with low overall flip rates or imbalanced group sizes. In such cases, even minor disparities could appear exaggerated, potentially misrepresenting the true fairness landscape. Practitioners should interpret these metrics with caution, considering contextual factors and employing additional statistical methods to validate and contextualize observed disparities.

The application of proportionality metrics should always be contextualized within specific normative frameworks relevant to the domain in question. For example, in healthcare, proportionality might entail accepting a certain level of disparity to prioritize the most urgent cases, while in employment or education, a more egalitarian proportionality might be ethically justified to correct historical inequities.

\section{Conclusions and Future Work}

This study explores the dynamics of bias mitigation within algorithmic decision-making systems, particularly emphasizing the unintended consequences arising from post-processing fairness interventions. We introduce a novel set of metrics explicitly designed to evaluate the proportionality of prediction flips resulting from these interventions. Furthermore, we propose an actionable methodology that integrates these proportionality metrics into existing machine learning workflows, enhancing transparency and accountability in algorithmic decisions.

Our contributions provide practitioners with tools for critically assessing whether fairness is equitably achieved across demographic groups, offering insights that extend beyond conventional fairness evaluations. These proportionality metrics serve as safeguards, promoting responsible and ethically justified deployments of algorithmic systems.

Future research directions include expanding and strengthening empirical evaluations. Specifically, we plan comprehensive experiments involving diverse real-world datasets, multiple classification models, and various post-processing fairness interventions to rigorously validate and generalize our metrics. Furthermore, exploring alternative normalization techniques (such as weighting proportionality metrics by group sizes or employing statistical validation methods) would further enhance the reliability of the metric. Extending our proportionality framework to multiclass classification settings and multiple protected attributes will also be crucial. Lastly, integrating these metrics into widely adopted fairness toolkits, such as AIF360 or Fairlearn, would significantly streamline fairness assessments, enabling practitioners to evaluate fairness, proportionality, and predictive accuracy within a unified analytical framework.

Additionally, we intend to extend our proportionality analysis by incorporating neighborhood-based individual metrics, enabling a detailed assessment of unintended consequences at the instance level and improving transparency and accountability in post-processing bias correction strategies.

\bibliography{main}

\appendix

\section{Source code}
The sources for this article are available via \href{https://github.com/juliettm/group-flip-proportionality}{GitHub}

\section{Summary of the proposed metrics}
\label{summary}

\begin{table}[h!]
\caption{Summary of the proposed metrics describing flips in the solution (can be used to describe different groups separately). Each line delineates the boundaries of the metric, includes a reference to their respective equations, and provides a short description along with any edge cases associated with them.}
\label{tab:sumary_solution}
\begin{tabular}{cccp{8cm}}
\hline
\textbf{Metric} & \textbf{Boundaries} & \textbf{Equation}  & \textbf{Short Description \& Edge Cases}\\
\midrule
FR & $[0, 1]$ & \ref{eq_fliprate} & Measures the proportion of instances where predictions change after debiasing. Is $0$ when there is no flip in the post-processing stage. Values close to $1$ suggest more flips. \\ 
DFR & $[0, \infty)$ & \ref{eq_dfr} & Ratio of beneficial to harmful flips. Indicates the balance between favorable and unfavorable flips. Values close to $1$ are desirable. Returns $\infty$ when there are no harmful flips. Returns $0$ when there are no beneficial flips and $1$ in the absence of flips.\\ 
HFP & $[0, 1]$ & \ref{eq_HFP} &  Proportion of flips leading to unfavorable outcomes. Values close to $1$ indicate a higher incidence of harmful flips. Is $0$ when there are no harmful flips. Take $1$ if all the flips present are harmful. \\
\bottomrule
\end{tabular}
\end{table}

\begin{table}[h!]
\caption{Summary of the proposed metrics to quantify the proportionality of the flips between groups. Each line delineates the boundaries of the metric, includes a reference to their respective equations, and provides a short description along with any edge cases associated with them.}
\label{tab:sumary_group}
\begin{tabular}{cccp{8cm}}
\hline
\textbf{Metric} & \textbf{Boundaries} & \textbf{Equation}  & \textbf{Short Description \& Edge Cases} \\
\midrule
FRD \& HFPD & $[0, 1]$ & \ref{eq_FRD} \& \ref{eq_HFPD} & Absolute differences in flips rates or harmful flips proportions between the groups. Take $0$ value when FR or HFP are equal in both groups. Values close to $0$ indicates greater proportionality. \\ 
DI \& HDI & $[1, \infty)$ & \ref{eq_DI} \& \ref{eq_HDI} & Ratio of flip rates between groups (DI). Ratio of harmful flip proportions between groups (HDI).  Returns $\infty$ when the minimum value of the flip rate or the harmful flip proportion is $0$. Returns $1$ when both values are equal or $0$. \\
FD \& HFD & $[0, \infty)$ & \ref{eq_FD} \&  \ref{eq_HFD} & Quantify the disparities in the flip rates or the harmful flip proportion between the groups, normalized by the overall flip rate. It has the potential to become significantly large when the overall flip rate or harmful flip proportion approaches $0$. Returns $1$ when both values are $0$ and $\infty$ when one of them is $0$. \\ 
RFD \& RHFD & $[0, 1]$ & \ref{eq_RFD} \& \ref{eq_RHFD} & Relative disparity in flip rates between groups (RFD). Relative disparity in harmful flips between groups (NHFD). It is a normalized measure of disparity, making it easier to compare in different scenarios. Returns $1$ when one rate is $0$ and the other is not, and returns $0$ when there are no flips in any of the groups. \\
\bottomrule
\end{tabular}
\end{table}

\end{document}